\documentclass[acmtog,screen]{acmart}

\usepackage{multirow}
\usepackage{amsmath}
\usepackage{colortbl}
\usepackage{bbding}
\usepackage{booktabs}
\usepackage{multirow}
\usepackage{siunitx}

    \newcommand{\edit}[1]{{{{#1}}}}

\copyrightyear{2026}
\acmYear{2026}
\setcopyright{cc}
\setcctype{by}
\acmConference[SIGGRAPH Conference Papers '26]{Special Interest Group on Computer Graphics and Interactive Techniques Conference Conference Papers}{July 19--23, 2026}{Los Angeles, CA, USA}
\acmBooktitle{Special Interest Group on Computer Graphics and Interactive Techniques Conference Conference Papers (SIGGRAPH Conference Papers '26), July 19--23, 2026, Los Angeles, CA, USA}
\acmDOI{10.1145/3799902.3811040}
\acmISBN{979-8-4007-2554-8/2026/07}

\begin{document}

\title{Autoregressive B-Rep Shape Generation with Parametric Surfaces}

\author{Dafei Qin}
\authornote{Equal contribution. ‡ Project lead. † Corresponding authors.}
\affiliation{\institution{The University of Hong Kong, Deemos Technology Co., Ltd.}
\country{China}}
\email{qindafei@connect.hku.hk}

\author{Rui Xu}
\authornotemark[1]
\affiliation{\institution{The University of Hong Kong, Deemos Technology Co., Ltd.}
\country{China}}
\email{ruixu1999@connect.hku.hk}

\author{Zeyu Shen}
\affiliation{\institution{Institute of Software, Chinese Academy of Sciences} 
\country{China}}
\email{shenzy@ios.ac.cn}

\author{Kaichun Qiao}
\affiliation{\institution{ShanghaiTech University, Deemos Technology Co., Ltd.} 
\country{China}}
\email{qiaokch2022@shanghaitech.edu.cn}

\author{Hongyang Lin}
\affiliation{\institution{ShanghaiTech University, Deemos Technology Co., Ltd.} 
\country{China}}
\email{linhy@shanghaitech.edu.cn} 

\author{Qixuan Zhang}
\authornotemark[3]
\affiliation{\institution{ShanghaiTech University, Deemos Technology Co., Ltd.} 
\country{China}}
\email{zhangqx1@shanghaitech.edu.cn}

\author{Huaijin Pi}
\affiliation{  \institution{The University of Hong Kong}
\country{China}}
\email{huaijinpi@connect.hku.hk}

\author{Lan Xu}
\authornotemark[2]
\affiliation{\institution{ShanghaiTech University} 
\country{China}}
\email{xulan1@shanghaitech.edu.cn}

\author{Jingyi Yu}
\affiliation{\institution{ShanghaiTech University} 
\country{China}}
\email{yujingyi@shanghaitech.edu.cn}

\author{Wenping Wang}
\affiliation{  \institution{Texas A\&M University}
\country{USA}}\email{wenping@tamu.edu}

\author{Taku Komura} 
\authornotemark[2]
\affiliation{  \institution{The University of Hong Kong}
\country{China}}
\email{taku@cs.hku.hk}

\begin{abstract}
Generative CAD modeling has broad design and application potential.
Despite significant advances in Boundary Representation (B-Rep) generation, the dominant representation in CAD, existing methods largely depend on uniformly sampled point- or grid-based geometry representations, sacrificing native surface types and parameters and thereby limiting geometric fidelity and downstream usability.
We present ParaCAD, an autoregressive framework for point-cloud-conditioned B-Rep generation that directly operates on native parametric surfaces.
ParaCAD introduces a surface-centric tokenization that explicitly encodes each face by its exact surface type and continuous parameters, preserving the intrinsic semantics of CAD geometry.
Our model first generates parametric surfaces with constrained UV domains, and then constructs a valid B-Rep by globally intersecting these surfaces to recover edges and vertices.
ParaCAD places point-cloud-conditioned generation at the core of B-Rep synthesis, making it practical for user-guided reconstruction and seamless integration into existing 3D generation pipelines.
Extensive experiments demonstrate that ParaCAD produces accurate B-Reps with faithful point-cloud alignment, outperforming point-based baselines in geometric precision, robustness, \edit{watertightness} and downstream usability.
 
\end{abstract}

\ccsdesc[500]{Computing methodologies~Parametric curve and surface models}

\begin{teaserfigure}
  \includegraphics[width=\textwidth]{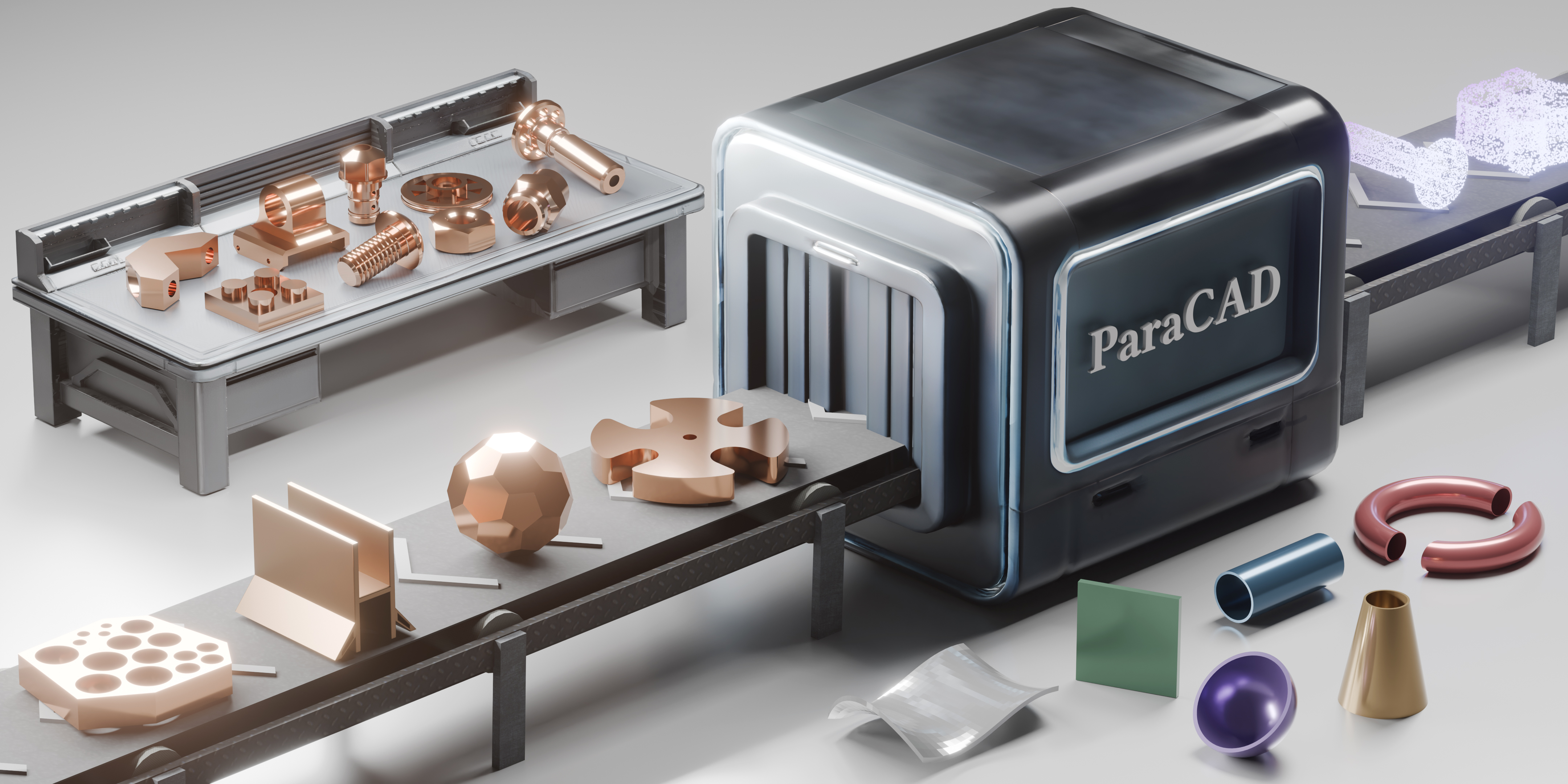}
  \vspace{-6mm}
  \caption{ParaCAD receives point clouds as conditions and generates CAD models with native support of original parametric surfaces.  }
  \label{fig:teaser}
\end{teaserfigure}

\maketitle

\section{Introduction}
\label{sec:intro}

Computer-Aided Design (CAD) serves as a critical bridge between conceptual design and physical realization by providing precise geometric descriptions.
Modern CAD systems predominantly represent solids using Boundary Representation (B-Rep)~\cite{weiler1986topological}, where geometry and topology are jointly defined by vertices, curves, and surfaces with exact analytical formulations.
This deliberate structural design ensures that every geometric element is explicitly parameterized and semantically meaningful, enabling robust downstream operations, including precise editing, constraint enforcement, simulation, and manufacturing.
As a result, preserving native surface types and their parametric definitions is fundamental to maintaining the fidelity, editability, and practical utility of CAD models.

\begin{figure}[!tp]
    \centering
\includegraphics[width=.8\linewidth]{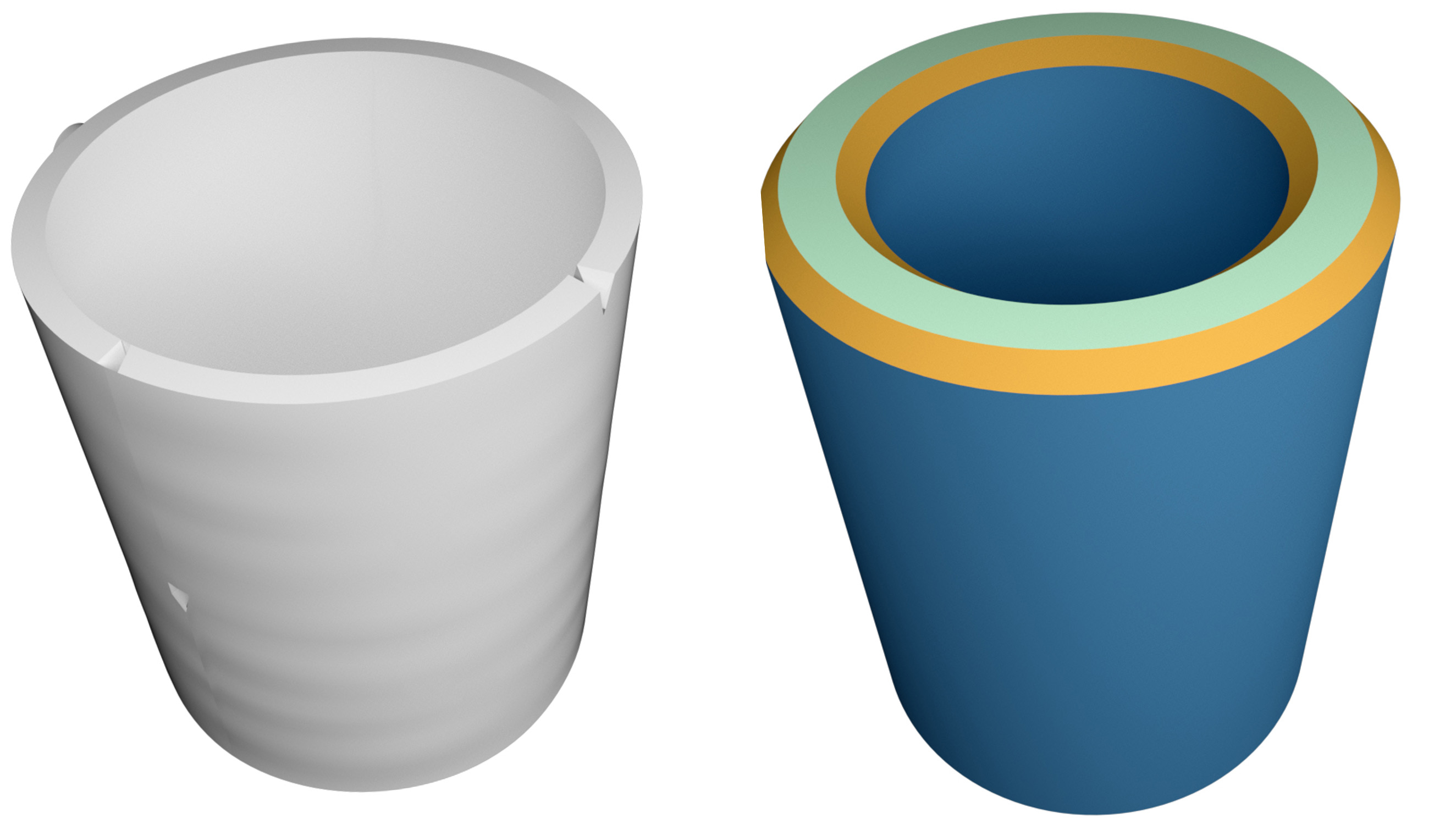}
    \caption{\textbf{Comparison between B-Reps representations} generated by (left) grid-based surface approximation and (right) native parametric surface representation.
    Surfaces are shaded to highlight different types.}
    \label{fig:grid-based-compare}
\end{figure}

Recent work on direct B-Rep generation~\cite{xu2024brepgen, li2025dtgbrepgen, liu2025hola, lee2025brepdiff, xu2025autobrep, li2025brepgpt} uses latent grid or point-based representations to encode each surface as a compact latent vector, providing a unified, network-friendly representation across diverse surface types.
However, these representations introduce compression error and blur surface-type semantics. Simple analytic surfaces may no longer be faithfully preserved. As shown in Fig.~\ref{fig:grid-based-compare}, B-Reps generated via latent grid-based representation (left) can not guarantee the accuracy of the enclosing cylindrical surface.
These errors make downstream geometric computations unreliable because operations such as surface–surface intersection can amplify small deviations. For example, Fig.~\ref{fig:grid-based-compare} shows that small holes may appear in their generated B-Reps (left) due to the errors amplified by the intersection of approximated representations. 
In addition, existing methods often avoid primitive recovery and instead represent everything as freeform geometry~\cite{xu2024brepgen, liu2025hola, xu2025autobrep}.
As a result, the generated B-Reps become harder to interpret and edit, which weakens key CAD advantages for manufacturing and simulation.

To mitigate these issues, we introduce a novel framework named \textbf{ParaCAD}, a GPT-based autoregressive generator designed to directly predict B-Rep surfaces with explicit UV parameterization.
Our core innovation is a hybrid surface tokenizer that encodes B-Rep faces in their native parametric forms.
Specifically, our tokenizer represents each surface instance with an explicit surface-type token. 
For common analytic primitives, this is followed by a compact set of continuous parameter tokens, while for general freeform patches, we utilize Finite Scalar Quantization (FSQ)~\cite{mentzer2023finite} to encode the geometry into discrete latent tokens. 
Conditioned on point clouds, ParaCAD produces B-Rep surfaces that align closely with the observed geometry while preserving surface types and parameters.

This design offers several practical advantages.
First, common analytic primitives are represented via explicit continuous parameters rather than approximate latent codes, ensuring that they can be accurately reconstructed. As shown in Fig.~\ref{fig:grid-based-compare}, our generated B-Reps (right) preserve the exact geometric smoothness.
Second, for complex freeform surfaces, we employ a compact latent representation that remains structurally consistent with the analytic tokens.
This unified formulation simplifies the output space, making autoregressive generation significantly more tractable and stable.
Third, because generated surfaces retain precise mathematical definitions, ParaCAD can robustly leverage standard geometric routines like surface intersections. 
Thus, we can deliberately defer topology construction to a deterministic geometric intersection stage, where edges and vertices are recovered by computing surface–surface intersections.
This reduces the amount of explicit structure that must be serialized into tokens, and hence shortens sequences and facilitates scaling autoregressive modeling to more complex solids.

To train ParaCAD, we construct a large-scale point cloud to CAD dataset of 583K samples derived from ABC~\cite{Koch_2019_CVPR}.
This dataset provides comprehensive resources for point-cloud-conditioned B-Rep generation.

We evaluate ParaCAD on this dataset against representative baselines for point-cloud-based B-Rep generation.
Results show that ParaCAD achieves substantially higher geometric accuracy and surface-type consistency. 
It also improves downstream CAD usability by producing outputs that better support editing operations and manufacturability.

In summary, our contributions are:
\begin{enumerate}
\item A \textbf{surface-centric tokenizer} that represents B-Rep faces as discrete tokens while preserving native surface types and parameters. It explicitly encodes common analytic primitives and supports freeform patches through a compatible discretization
\item An \textbf{effective point-cloud-conditioned autoregressive generator} that predicts parametric surfaces with bounded UV domains. By removing explicit topological entities from the learning space, we substantially simplify the generative task, allowing the model to focus on predicting geometrically meaningful surfaces
\item An \textbf{exact geometric solver} that robustly assembles the B-Rep from parametric surfaces. 
By leveraging surface-surface intersection and binary integer programming, we cut surface patches into pieces and assemble useful pieces to a closed manifold, maximizing geometric coverage while ensuring topological validity.
\end{enumerate}

\section{Related Works}
\label{sec:related}
\subsection{CAD Generation via Construction Operation}
A substantial body of learning-based CAD generation work~\cite{para2021sketchgen, ganin2021computer,willis2021engineering, wu2021deepcad, xu2022skexgen, seff2022vitruvion, xu2023hnc,guo2025cadtrans} represent CAD as a sequence of high-level construction operations (e.g., sketch, extrude, revolve) or Boolean programs.
This paradigm has been extended to conditional generation from diverse modalities, including sketches~\cite{li2020sketch2cad, para2021sketchgen}, images~\cite{alam2025gencad,chen2025img2cad,you2025imagetocad}, voxels~\cite{lambourne2022reconstructing}, \edit{point-cloud~\cite{rukhovich2025cad}, B-Reps~\cite{Xu_2021_CVPR}} and language~\cite{khan2024text2cad,xu2024CADMLLM}.
However, sequential command-based models are constrained by limited set of supported commands and their reliance on construction histories significantly increases data collection costs, and existing datasets for such sequences remain substantially smaller than those available for B-Rep models~\cite{Koch_2019_CVPR}.

\subsection{Constructive Solid Geometry Generation}
In parallel, constructive solid geometry (CSG)~\cite{requicha1977construcive} models represent shapes as compositions of simple primitives with Boolean operations, and have been studied under both supervised and unsupervised learning setups. Early works relied on program synthesis to reconstruct CSG representations~\cite{du2018inversecsg,nandi2017programming,nandi2018functional}, while later learning-based approaches improved robustness and scalability through neural program induction and parsing~\cite{sharma2018csgnet,ellis2019write, kania2020ucsg, yu2022capri,yu2023d, tian2018learning, chen2020bsp}.
Despite these advances, CSG-based methods face inherent ambiguity, as multiple CSG sequences can lead to the same result. Besides, the limited primitive vocabulary and dataset restrict the scalability of these methods.

\subsection{Direct B-Rep Generation}
\subsubsection{Unconditional Generation}
Learning-based CAD modeling has increasingly focused on boundary representation (B-Rep)~\cite{lee2001partial} generation, where the core challenge lies in jointly modeling heterogeneous parametric primitives and their discrete topological relations~\cite{lambourne2021brepnet, jones2023self, jayaraman2021uv, xu2022rfeps, xu2024cwf, dong2024neural}. UV-Net~\cite{jayaraman2021uv} popularized a unified geometry encoding by sampling curves and surfaces on regular grids in their parametric domains.
Following this grid-based paradigm, generative methods have made rapid progress in direct B-Rep generation~\cite{jayaraman2023solidgen,guo2022complexgen, xu2024brepgen, li2025dtgbrepgen, fan2024neuronurbs}. For example,  BrepGen performs hierarchical diffusion over structured latent geometry, generating entities with strong inter-stage dependencies and relying on post-processing (e.g., de-duplication/merging) for topological consistency.
DTGBrepGen~\cite{li2025dtgbrepgen} stresses validity by explicitly decoupling topology and geometry, improving structural consistency at the cost of a more staged generation process.
NeuroNURBS~\cite{fan2024neuronurbs} explores B-spline primitive-based encodings to better preserve parametric structure.
More recently, HoLa~\cite{liu2025hola} simplifies the pipeline by learning a holistic latent space defined primarily on surfaces and inferring curves/connectivity implicitly via learned surface--surface intersection modules.
Recent AutoRegressive (AR) approaches instead cast B-Reps as sequence modeling problems.
AutoBrep~\cite{xu2025autobrep} proposes a unified tokenization that interleaves geometric latent tokens with topological reference tokens and uses a face-adjacency traversal for serialization, enabling transformer-based generation and autocompletion.
BrepGPT~\cite{li2025brepgpt} similarly adopts a decoder-only AR framework with vector-quantized tokens over local geometric--topological units.
Notably, despite being autoregressive and more unified than cascaded pipelines, these AR formulations still explicitly account for B-Rep combinatorics through dedicated topology tokens or local connectivity units, such as edges and vertices.

On the other hand, BrepDiff~\cite{lee2025brepdiff} proposes a single-stage diffusion model directly over surface-level UV-grid representations, completely avoiding generation of explicit vertices, edges, or topology graphs thus greatly simplifies the generation pipeline. However, they still rely on the grid-based approximation protocol to represent surfaces, thus their network outputs lose primitive types and precise parameterized representations. In contrast, we keep the same surface-level simplification but directly tokenize and generate native parametric surface types and parameters, preserving explicit CAD semantics without grid-based approximations.

\subsubsection{Point-Cloud Conditioned Reconstruction and Generation}

Existing point-cloud conditioned B-Rep generation methods largely follow either segmentation and fitting pipelines or direct structured prediction with intermediate representations and features. 
ParSeNet~\cite{sharma2020parsenet} decomposes point clouds into parametric surface patches, providing editable and interpretable parametrization. 
HPNet~\cite{yan2021hpnet} and SEDNet~\cite{li2023surface} segment point clouds into primitive or surface regions that are subsequently converted into CAD entities via downstream fitting and assembly.
Point2CAD~\cite{liu2024point2cad} builds a pipeline that combines face clustering with analytic or learned surface fitting and subsequent topology recovery. Split-and-Fit~\cite{liu2024split} learns Voronoi partition of point-cloud to better segment the conditional input.  Despite their differences, these methods rely on primitive detection fitting to obtain a valid B-Rep, and therefore are limited by the primitive fitting algorithm and struggle to support diverse freeform surfaces.

On the direct side, ComplexGen~\cite{guo2022complexgen} introduces a chain-complex representation to jointly predict geometric primitives and their incidence relations from point clouds, enabling structured generation of B-Rep elements. 
Recent learning-based methods~\cite{li2025dtgbrepgen, liu2025hola, li2025brepgpt} perform conditioned generation by injecting point-cloud information from pretrained VAEs ~\cite{zhao2023michelangelo}, yet their results are mediated by learned latent representations rather than directly emitting native parametric surface types and parameters. 
In contrast, our method conditions generation on point clouds while directly tokenizing and generating explicit parametric surface types and parameters, and also naturally supporting freeform surfaces and curves without primitive fitting.

\section{Methods}
\label{sec:methods}
\begin{figure}[!tp]
    \centering
\includegraphics[width=\linewidth]{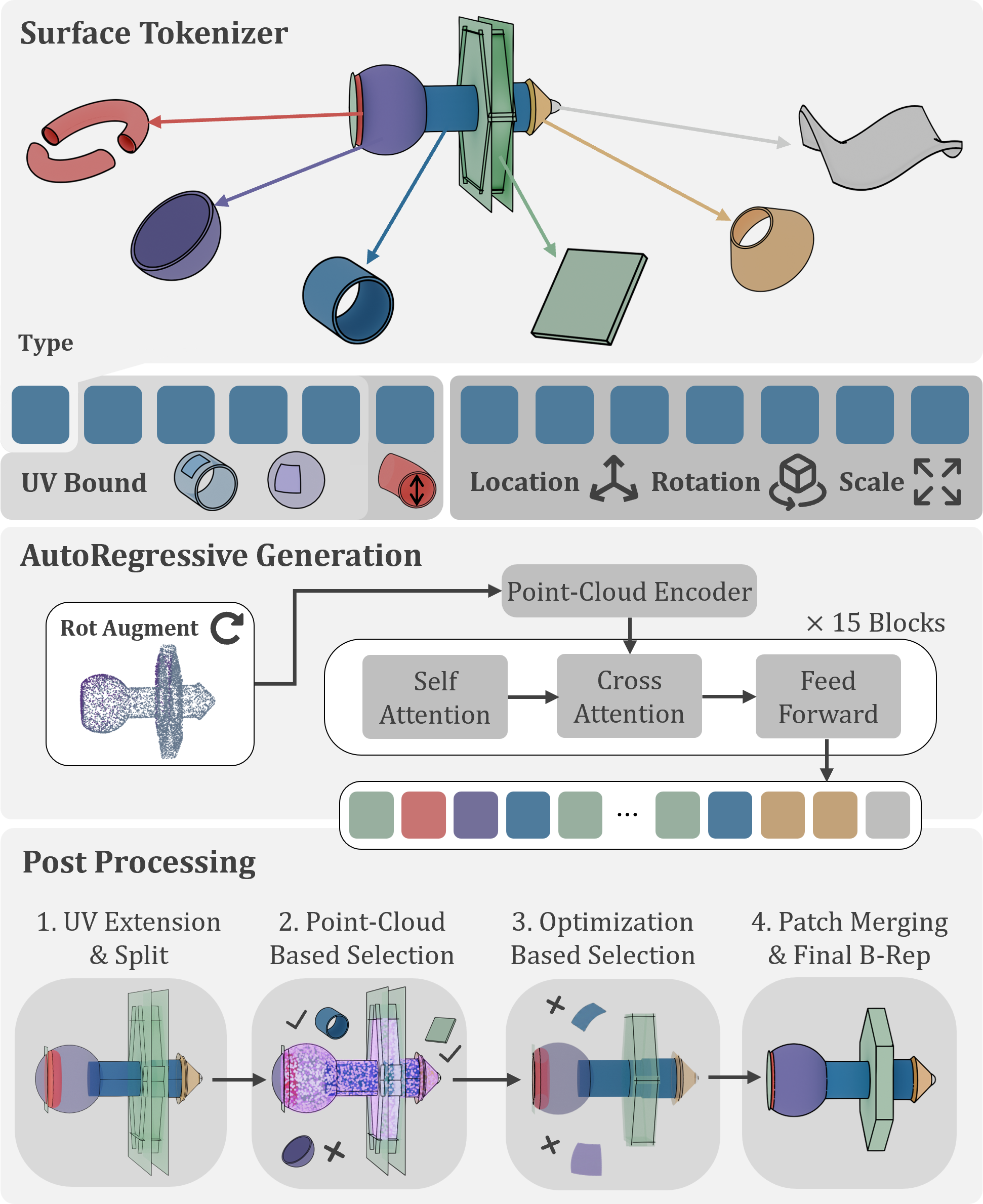}
    \caption{\textbf{ParaCAD Pipeline.} (Top) The surface tokenizer records and discretizes the parametric surface representation. (Middle) The decoder-only autoregressive model generates sequences of surface tokens from point-cloud conditions. (Bottom) Geometrical operations for consolidating the final B-Rep.}
    \label{fig:pipeline}

\end{figure}
Given a 3D point cloud, our goal is to generate a CAD model, where the boundary representation (B-Rep) format is employed to describe the geometry and topology.
To achieve this, we propose ParaCAD, a novel framework that directly predicts B-Rep surfaces with UV parameterization.
As shown in Fig.~\ref{fig:pipeline}, ParaCAD consists of three components.

First, in Sec.~\ref{sec:tokenizer} we introduce a surface tokenizer that maps parametric surface descriptions into a compact discrete space.
Second, in Sec.~\ref{sec:ar} we describe the point-cloud-conditioned autoregressive generator. It produces a sequence of surface patches with explicit UV domains following the geometry of the point cloud.
Third, we discuss in Sec.~\ref{sec:post} a geometric post-processing stage that performs global surface intersections, clipping, and sewing to assemble a final B-Rep, using the point-cloud condition to suppress ambiguities during construction.

\subsection{Surface Tokenizer}
\label{sec:tokenizer}

In ParaCAD, each constituent face is parameterized by an underlying (generally unbounded) analytic surface type together with an explicit rectangular parameter range.
Specifically, we represent the $i$-th face by a native parametric surface with explicit intrinsic and extrinsic parameters.

Formally, let $\tau_i \in \mathcal{T}$ denote the surface type of the $i$-th face (e.g., plane, cylinder, cone, sphere, torus, and freeform).
We define the corresponding surface patch as the restriction of the type-specific parametric mapping
\begin{equation}
S_{\tau_i}(u,v;\boldsymbol{\beta}_i,\boldsymbol{\pi}_i):
[u_{\min}^i,u_{\max}^i] \times [v_{\min}^i,v_{\max}^i]
\rightarrow \mathbb{R}^3,
\label{eq:surface_def_main}
\end{equation}
where $\boldsymbol{\beta}_i$ denotes the intrinsic (shape) parameters of the surface family $\tau_i$ (e.g., radius or semi-angle),
and $\boldsymbol{\pi}_i$ encodes the extrinsic placement (location and orientation) in $\mathbb{R}^3$.
The closed intervals $[u_{\min}^i,u_{\max}^i]$ and $[v_{\min}^i,v_{\max}^i]$ specify the rectangular UV range considered for this face.
The analytic forms of $S_{\tau}$ for each surface family in $\mathcal{T}$ are provided in the appendix.

\paragraph{Token block.}
Our surface tokenizer serializes each bounded parametric surface patch into a fixed-length block of 13 discrete tokens.
Let $\mathbf{s} = \big[s_0, s_1, \dots, s_{12}\big] \in \mathbb{Z}^{13}$ denote the token block for one face.
Token $s_0$ encodes the surface type $\tau$.
Tokens $s_1$ to $s_5$ encode intrinsic surface information in a canonical coordinate system.
Tokens $s_6$ to $s_{12}$ encode an instance-level isotropic similarity transform ($\mathrm{SIM}$(3)) that maps the canonical surface to its placement in the original B-Rep.

\subsubsection{Intrinsic parameter tokenization}
\label{sec:tokenizer:intrinsic}

For the five elementary surface families we consider, i.e., plane, cylinder, cone, torus and sphere, we encode intrinsic geometry in canonical space using a compact, fixed-format tokenization.
Following ISO STEP B-Rep~\cite{ISO10303-42}, we define a \emph{canonical} surface as one whose 
local coordinate system has its location at the origin and its orthonormal frame aligns with the canonical axes $(\mathbf{e}_x,\mathbf{e}_y,\mathbf{e}_z)$.
We additionally apply a type-specific normalization that sets the primary scale parameter of the surface to unity, such as unit extent for planes and unit radius for cylinders, spheres, and tori. Please refer to the appendix for details.
Under this convention, the intrinsic parameters of a surface record only the remaining shape-defining attributes that cannot be removed by a $\mathrm{SIM}$(3) transform, including the bounded $uv$ domain and type-specific parameters such as the cone semi-angle.

\paragraph{Bounded $uv$ domain.}
We discretize the four uv bounds (Eq.~\ref{eq:surface_def_main}) into tokens $s_1$ to $s_4$ using uniform codebooks defined over fixed ranges.

\paragraph{Type-specific scalar parameters.}
We reserve token $s_5$ for cone and torus, as they need additional scalars to represent the shape:
\begin{equation}
s_5 =
\begin{cases}
\texttt{Quantize}(\theta) & \text{if } \tau=\texttt{cone}, \\
\texttt{Quantize}(r) & \text{if } \tau=\texttt{torus}, \\
\texttt{PAD}_{\tau} & \text{otherwise},
\end{cases}
\end{equation}
where $\theta$ denotes the cone semi-angle and $r$ denotes the torus minor radius.
Surfaces without additional intrinsic scalars set $s_5$ to a type-specific padding token $\texttt{PAD}_{\tau}$.

Overall, the intrinsic tokenization consists of four $uv$ tokens and up to one scalar token, resulting in an intrinsic length of $4{+}1$ tokens.

\subsubsection{$\mathrm{SIM}$(3) Tokenization}
\label{sec:tokenizer:canonical}

Given an analytic surface with placement location $\mathbf{c}\in\mathbb{R}^3$ and an orthonormal frame $(\mathbf{x},\mathbf{y},\mathbf{z})$, where $\mathbf{z}$ is the surface normal (or axis direction) and $\mathbf{x}$ defines the reference $u$ direction, we map the surface into the canonical coordinate system defined in Sec.~\ref{sec:tokenizer:intrinsic} using a $\mathrm{SIM}$(3) transform, which is composed of translation, rotation, and isotropic scale. 
Specifically, we translate $\mathbf{c}$ to the origin, rotate $(\mathbf{x},\mathbf{y},\mathbf{z})$ onto $(\mathbf{e}_x,\mathbf{e}_y,\mathbf{e}_z)$, and apply a type-specific scale $\alpha$:
\begin{equation}
\bar{\mathbf{p}} \;=\; \alpha\,\mathbf{R}\,(\mathbf{p}-\mathbf{c}),
\label{eq:canon_step}
\end{equation}
where $\mathbf{R}\in\mathrm{SO}(3)$ is chosen to satisfy $\mathbf{R}\mathbf{z}=\mathbf{e}_z$ and $\mathbf{R}\mathbf{x}=\mathbf{e}_x$.
We then discretize the instance-level transform parameters $(\mathbf{c},\mathbf{R},\alpha)$ into tokens $s_6$ to $s_{12}$.

To be more specific, for \emph{translation}, we quantize the three components of $\mathbf{c}=(c_x,c_y,c_z)$ independently to obtain tokens $s_6$ to $s_8$. For \emph{rotation}, we convert $\mathbf{R}$ to Euler angles $\boldsymbol{\rho}=(\rho_1,\rho_2,\rho_3)$ under a fixed convention and uniformly quantize each angle to obtain tokens $s_{9}$ to $s_{11}$. We represent the isotropic \emph{scale} by its logarithm and quantize $\log(\alpha)$ to obtain token $s_{12}$, which yields a more balanced resolution across multiplicative scale changes.

\paragraph{Handling out-of-range surfaces.} 
A small fraction of surfaces may fall outside the supported quantization ranges, which typically corresponds to near-degenerate cases such as extremely low-curvature cylinders or cones with very large radius.
For these surfaces, we approximate the surface as a freeform patch and discretize it using the freeform representation described in the following section.

\subsubsection{Freeform surface discretization}
\label{sec:tokenizer:freeform}
We follow the discrete latent grid representation~\cite{xu2025autobrep} to sample each freeform face on a regular $32\times32$ grid over the bounded UV domain, yielding a point tensor $\mathbf{G}\in\mathbb{R}^{32\times32\times3}$.
A convolutional autoencoder then compresses $\mathbf{G}$ into a compact latent grid $\mathbf{Z} \;=\; E(\mathbf{G}) \in \mathbb{R}^{4\times4\times3}$.

Then, we apply finite-scalar quantization~\cite{mentzer2023finite} to discretize the continuous latent grid $\mathbf{Z}$ into a small set of quantized latents $\hat{\mathbf{Z}}$. The VAE decoder then reconstructs $\hat{\mathbf{G}}=D(\hat{\mathbf{Z}})$ with a standard reconstruction objective.

We then serialize the quantized latents into a fixed set of 12 tokens.
These tokens populate the slots of our face block for freeform surfaces, namely $s_1$ through $s_{12}$.

Together, our tokenizer provides a compact and semantically grounded discrete representation that is directly compatible with autoregressive generation.

\subsection{AutoRegressive Generation}
\label{sec:ar}

Given an input point cloud $P=\{\mathbf{p}_i\}_{i=1}^{N}$, our goal is to synthesize a CAD model in B-Rep form by generating a sequence of discrete surface tokens.
Using the tokenizer in Sec.~\ref{sec:tokenizer}, each B-Rep face is mapped to a fixed-length token block, and an entire solid is represented as a single token sequence, $\mathbf{s} = (s_1, s_2, \dots, s_T), s_t \in \mathcal{V},$ where $\mathcal{V}$ is the token vocabulary and $T$ depends on the number of faces. To obtain a deterministic sequence order, we sort all faces by their placement location with lexicographic priority on the $z$, $x$, and $y$ coordinates, and then tokenize them in this order. We then wrap the token sequence with a \texttt{[start]} and an \texttt{[end]} token. 

We learn a conditional autoregressive model parameterized by $\gamma$ that factorizes the likelihood of the token sequence given the point cloud:
\begin{equation}
p_{\gamma}(\mathbf{s}\mid P) \;=\; \prod_{t=1}^{T} p_{\gamma}\!\left(s_t \mid s_{<t}, P\right).
\label{eq:ar_factor}
\end{equation}
Training is performed by minimizing the negative log-likelihood, which is implemented as the following token-level cross-entropy loss using teacher forcing:
\begin{equation}
\mathcal{L}_{\mathrm{CE}} \;=\; - \sum_{t=1}^{T} \log p_{\gamma}\!\left(s_t^{\ast} \mid s_{<t}^{\ast}, P\right),
\label{eq:ce}
\end{equation}
where $\mathbf{s}^{\ast}$ denotes the ground-truth token sequence.
At inference time, given a point-cloud condition, we sample tokens sequentially from Eq.~\eqref{eq:ar_factor} until an \texttt{[end]} token is produced, and then decode the generated tokens into parametric surfaces using the inverse tokenizer.

\subsection{Post-Processing}
\label{sec:post}

\begin{figure}[!htbp]
\centering
\includegraphics[width=0.9\linewidth]{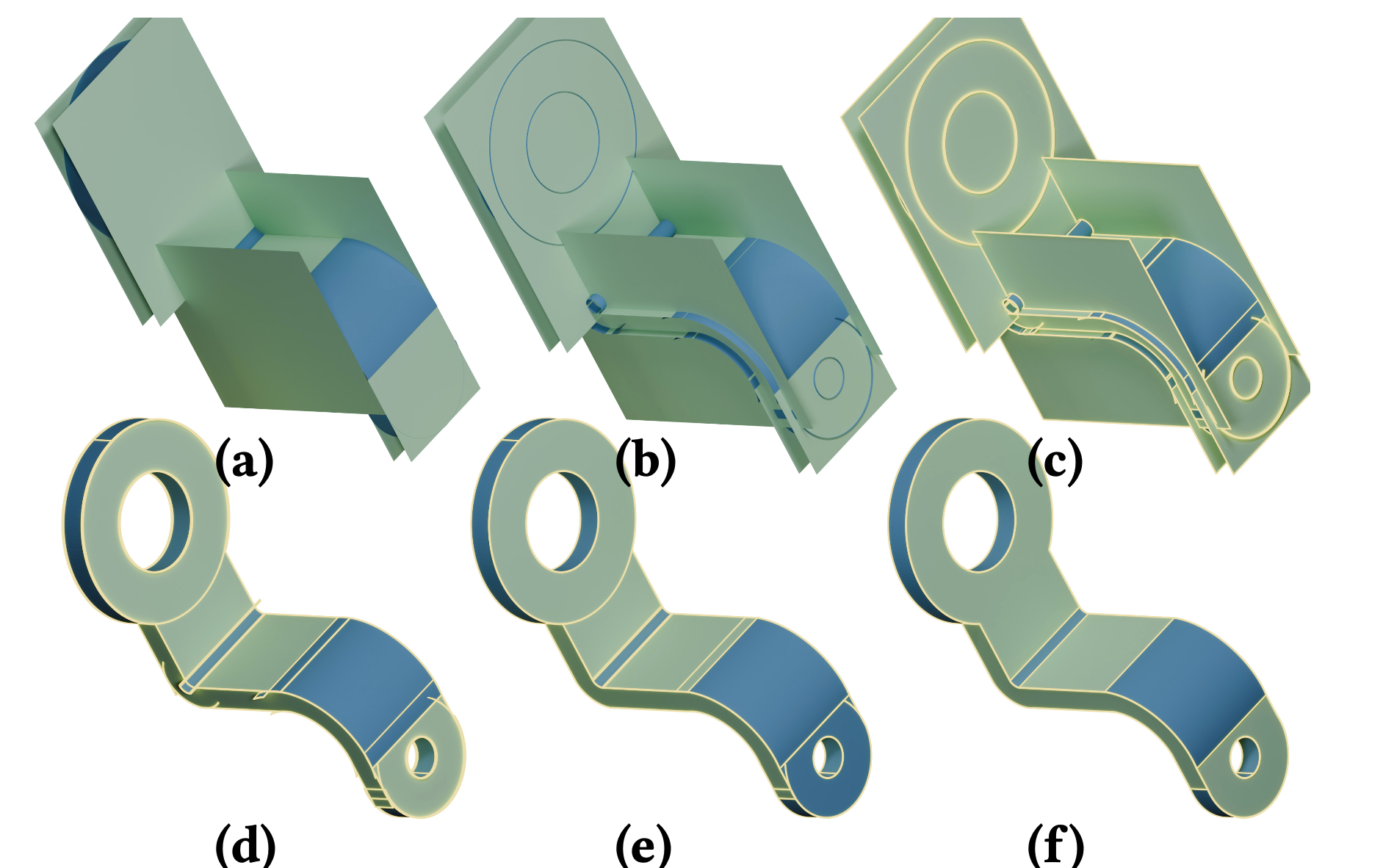}
        \caption{\textbf{Post-processing pipeline}. \textbf{(a)} Surface patches translated from generated tokens. \textbf{(b)} Surface UV extension. \textbf{(c)} Surface-surface intersection and split. \textbf{(d)} Point cloud coverage based selection. \textbf{(e)} Optimization-based selection. \textbf{(f)} Merged.}
        \label{fig:post}
\end{figure}

Given the generated surface patches and the input point cloud, we perform a split-and-selection method to obtain the final B-Rep, illustrated in Fig.~\ref{fig:post}. 

\paragraph{UV extension and surface split.} 
We recover the B-Rep topology by computing geometric intersections between the generated surface patches.
For robust intersection, we extend the UV domain of each surface to create sufficient overlap.
Thanks to the native parametric surface representation, we can perform surface extension exactly on the uv space, rather than interpolating the surface geometry~\cite{lee2025brepdiff, li2025caddreamer}.
Besides, we follow~\cite{shen2025mesh2brep} to detect tangent relations between adjacent surfaces, and adjust their parameters to ensure the existence of an intersection. We then perform surface-surface intersection and split surfaces into small pieces, as shown in Fig.~\ref{fig:post} (b).

\paragraph{Point-cloud based pieces selection.}
Given an input point cloud $P=\{\mathbf{p}\}\subset\mathbb{R}^3$, we select a subset of the surface pieces that best align with the local point distribution.
Let $\mathcal{S}=\{S_i\}$ denote the set of surface pieces.
For each piece $S_i$, we define a point coverage score $\mathrm{Cov}(S_i,P)\in[0,1]$ that measures how well $S_i$ is supported by $P$.

We uniformly sample a set of query points $Q_i=\{\mathbf{q}\}$ on the patch $S_i$ and evaluate whether each sample is supported by the point cloud in both distance and normal agreement:
\begin{equation}
\label{eqn:cov}
\mathrm{Cov}(S_i,P)
=
\frac{1}{|Q_i|}
\sum_{\mathbf{q}\in Q_i}
\mathbb{I}\!\Big(
\exists\,\mathbf{p}\in P:\;
\|\mathbf{q}-\mathbf{p}\| < \delta_d
\;\wedge\;
\big|\mathbf{n}_{\mathbf{q}}^\top \mathbf{n}_{\mathbf{p}}\big| > \delta_n
\Big),
\end{equation}
where $\mathbb{I}(\cdot)$ is the indicator function, $\mathbf{n}_{\mathbf{q}}$ and $\mathbf{n}_{\mathbf{p}}$ are unit normals at $\mathbf{q}$ and $\mathbf{p}$, respectively,
$\delta_d>0$ is a distance tolerance, and $\delta_n\in(0,1]$ is a normal-consistency threshold.
In our implementation, we set $\delta_n=\cos(\pi/6)$.

We apply a relative selection criterion to adapt to varying point densities.
A patch $S_i$ is discarded if its coverage is substantially below the best-supported candidate:
\begin{equation}
\label{eqn:cov_filter}
\mathrm{Cov}(S_i,P) < \lambda \cdot \max_{S_k\in\mathcal{S}} \mathrm{Cov}(S_k,P),
\end{equation}
where $\lambda\in(0,1)$ is a relaxation factor (we use $\lambda=0.8$). The selected pieces are shown in Fig.~\ref{fig:post} (c).

\paragraph{Optimization-based pieces selection.}
To further filter candidate pieces, we solve a binary program that maximizes coverage while enforcing topological consistency between selected faces and edges.

Let $\mathcal{F}=\{f_i\}$ be the set of candidate pieces and $\mathcal{E}=\{e_j\}$ the set of candidate edges.
We introduce binary indicators
$
x_i\in\{0,1\}
$
for selecting face $f_i$, and
$
y_j\in\{0,1\}
$
for selecting edge $e_j$.
Each face $f_i$ is associated with a coverage score $\mathrm{Cov}_i\in\mathbb{R}_{\ge 0}$ defined in Eq.~\ref{eqn:cov}.
We partition edges into three categories:
inner edges $\mathcal{E}^{\mathrm{i}}$ (adjacent to exactly two candidate faces),
boundary edges $\mathcal{E}^{\mathrm{b}}$ (adjacent to exactly one candidate face),
and conflict edges $\mathcal{E}^{\mathrm{c}}$ (adjacent to more than two candidate faces, arising from surface splits).
For an edge $e_j$, let
$
\mathrm{NF}(e_j)\subseteq \mathcal{F}
$
denote the set of incident faces. We reward high-coverage faces, encourage selecting inner edges, and penalize boundary edges with weight $w_{\mathrm{b}}>0$:
\begin{align}
\max_{\mathbf{x},\mathbf{y}}\quad
& \sum_{f_i\in\mathcal{F}} x_i\,\mathrm{Cov}_i
\;+\; \sum_{e_j\in\mathcal{E}^{\mathrm{i}}\cup\mathcal{E}^{\mathrm{c}}} y_j
\;-\; w_{\mathrm{b}}\sum_{e_j\in\mathcal{E}^{\mathrm{b}}} y_j .
\label{eq:patch_obj}
\end{align}
For inner and conflict edges, an edge is activated if and only if exactly two of its incident faces are selected, enforcing the 2-manifold property:
\begin{align}
\sum_{f_i\in \mathrm{NF}(e_j)} x_i \;-\; 2\,y_j \;=\; 0,
\qquad \forall\, e_j\in\mathcal{E}^{\mathrm{i}}\cup\mathcal{E}^{\mathrm{c}}.
\label{eq:patch_cons_inner}
\end{align}
For boundary edges, $y_j$ directly tracks whether the sole incident face is selected:
\begin{align}
x_i \;-\; y_j \;=\; 0,
\qquad \forall\, e_j\in\mathcal{E}^{\mathrm{b}},\; f_i\in\mathrm{NF}(e_j).
\label{eq:patch_cons_border}
\end{align}
This formulation encourages closed, watertight shells: each selected boundary edge incurs a penalty of $w_{\mathrm{b}}$, discouraging open boundaries unless justified by high coverage scores.

\paragraph{Patches merging.} Finally, we merge the selected pieces based on their original piece affiliation to form the final B-Rep solid (Fig.~\ref{fig:post} (f)).

The reliability of the post-processing algorithm is supported by the following two features. First, the use of exact parametric primitives reduces per-face approximation error, making intersection and sewing more efficient and stable than grid-compressed alternatives.
Second, point cloud conditioning suppresses ambiguity by anchoring the solution to the observed shape.

\section{Experiments}
\label{sec:exp}
\subsection{Dataset construction}
\label{sec:data}

We construct a large-scale point-cloud-to-CAD dataset from the ABC dataset~\cite{Koch_2019_CVPR}.
For each B-Rep model, we first decompose it into its smallest connected components and scale each component into a unit box.
To reduce duplicates, we follow the filtering protocol of~\cite{xu2025autobrep} and remove repeated parts using a hash computed from the part name, edge and vertex counts, and the distribution of primitive surface types (e.g., cones and cylinders). This yields a total of 583K B-Rep instances.

For each B-Rep solid, we generate a high-fidelity conditioning point cloud by sampling points directly on the exact CAD geometry using OpenCASCADE~\cite{opencascade}.
We sample each face within its valid parametric domain with a fixed number of points, and then apply farthest point sampling to obtain a uniformly distributed point set of 10{,}240 points.  We split the resulting dataset into 90\% training, 5\% validation, and 5\% test sets.

\subsection{Implementation Details}
\label{sec:train}

ParaCAD adopts a decoder-only GPT with 15 Transformer blocks, with 768-dim embeddings and 12 heads, a 1,024-size codebook, and a 1,003-token context window that supports up to 77 surface patches. We condition generation on a point cloud using 257 tokens from a point-cloud encoder~\cite{zhao2023michelangelo}, and inject this conditioning via cross-attention in every Transformer block.

During training, we randomly subsample 4,096 points from each 10,240-point cloud and add isotropic Gaussian noise ($\sigma=0.005$). We further apply random right-angle rotations around the $x$, $y$, or $z$ axis with angles in ${0^\circ,90^\circ,180^\circ,270^\circ}$. Training runs for 5 days on 16 A800 GPUs.

\subsection{Results}

\subsubsection{Point-Cloud Conditioned B-Rep Generation}

We compare ParaCAD against point-cloud-conditioned B-Rep generation baselines, including ParseNet~\cite{sharma2020parsenet} / \edit{SEDNet~\cite{li2023surface}} + Point2CAD~\cite{liu2024point2cad} (P2CAD; segmentation then reconstruction), HoLa~\cite{liu2025hola} (latent diffusion generation), and NVDNet~\cite{liu2024split} (voronoi segmentation generation). All methods are evaluated on 1,000 randomly sampled test shapes from our test set. Following their standard settings, P2CAD and NVDNet take the full 10,240-point input, while HoLa and ParaCAD use 4,096 randomly sampled points. We use the official \edit{model and processing code} released by each baseline. \edit{For generative methods like HoLa and ParaCAD, we select best-of-16 and -4 for each case, respectively.} 

\paragraph{Geometry Similarity.} We calculate the Chamfer Distance (CD) between the generated point cloud and the ground truth point cloud. For P2CAD and NVDNet, the point clouds were the vertices of their output meshes. For HoLa and our method, we sampled 4,096 points from the generated B-Rep.

\paragraph{Semantic Fidelity.} We use type-conditioned surface coverage metrics to jointly capture geometric alignment and surface-type correctness. Specifically, we report Surface Precision (SP), Surface Recall (SR), and their harmonic mean (F1). SP measures the fraction of predicted surface area supported by same-type ground-truth faces, penalizing over-generation and type confusion, while SR measures the fraction of ground-truth surface area covered by same-type predictions, capturing missing or misclassified surfaces. Formal definitions and implementation details are provided in the appendix.

\begin{table}[h]
  \centering
  \small
  \setlength{\tabcolsep}{6pt}
  \renewcommand{\arraystretch}{1.15}
  \caption{\textbf{Quantitative comparison} on geometry similarity (CD) and semantic fidelity (SP/SR/F1). Note that HoLa generates all surfaces as freeform, thus fidelity is not applicable.}
  \label{tab:quantitative}
  \begin{tabular}{l c c c c}
    \toprule
    \multirow{2}{*}{Method} &
    \multicolumn{1}{c}{Geometry} &
    \multicolumn{3}{c}{Fidelity} \\
    \cmidrule(lr){2-2} \cmidrule(lr){3-5} 
    & CD ($\times 10^{-2}$) $\downarrow$ & SP $\uparrow$ & SR $\uparrow$ & F1 $\uparrow$ \\
    \midrule
    P2CAD (ParseNet)  & 4.266 & 0.133 & 0.314 & 0.177  \\
    P2CAD (SEDNet)  & 1.470 & 0.091 & 0.247 & 0.125  \\
    HoLa               & 3.650 & -- & -- & --  \\
    NVDNet             & 1.207 & 0.475 & 0.592 & 0.502  \\
    \midrule
    \textbf{Ours}      & \textbf{1.140} & \textbf{0.797} & \textbf{0.776} & \textbf{0.784}  \\
    \bottomrule
  \end{tabular}

\end{table}

As shown in Tab.~\ref{tab:quantitative}, our method consistently outperforms all baselines in both geometric similarity and semantic fidelity. The performance of P2CAD is constrained by segmentation quality and limited surface-type support; HoLa shows weaker adherence to the input point cloud due to its unconditional generation training scheme; and NVDNet lacks freeform surface modeling, reducing fit quality on shapes requiring such patches. Our explicit coverage of the five canonical analytic surface types and freeform surfaces, together with pre-trained priors, provides a clear advantage. Per-surface-type F1 results are reported in Tab.~\ref{tab:surface_type_f1}.
\begin{table}[h]
  \centering
  \small
  \setlength{\tabcolsep}{6pt}
  \renewcommand{\arraystretch}{1.2}
    \caption{\textbf{Per-type surface classification F1 scores}. Unsupported types are marked with `--'.}

  \label{tab:surface_type_f1}
  \resizebox{1.0\columnwidth}{!}{
  \begin{tabular}{l|ccccccc}
    \toprule
    Method &
    Plane &
    Cylinder &
    Cone &
    Sphere &
    Torus &
    Freeform \\
    \midrule
    P2CAD
      & 0.3053
      & 0.1960
      & 0.0643
      & 0.0120
      & --
      & -- \\
    NVDNet
      & 0.5893
      & 0.3968
      & 0.4566
      & 0.0390
      & 0.0458
      & -- \\
    Ours
      & \textbf{0.8229}
      & \textbf{0.6508}
      & \textbf{0.7189}
      & \textbf{0.6799}
      & \textbf{0.6082}
      & \textbf{0.4870} \\
    \bottomrule
  \end{tabular}
  }

\end{table}

\begin{figure}[!tp]
    \centering
\includegraphics[width=\linewidth]{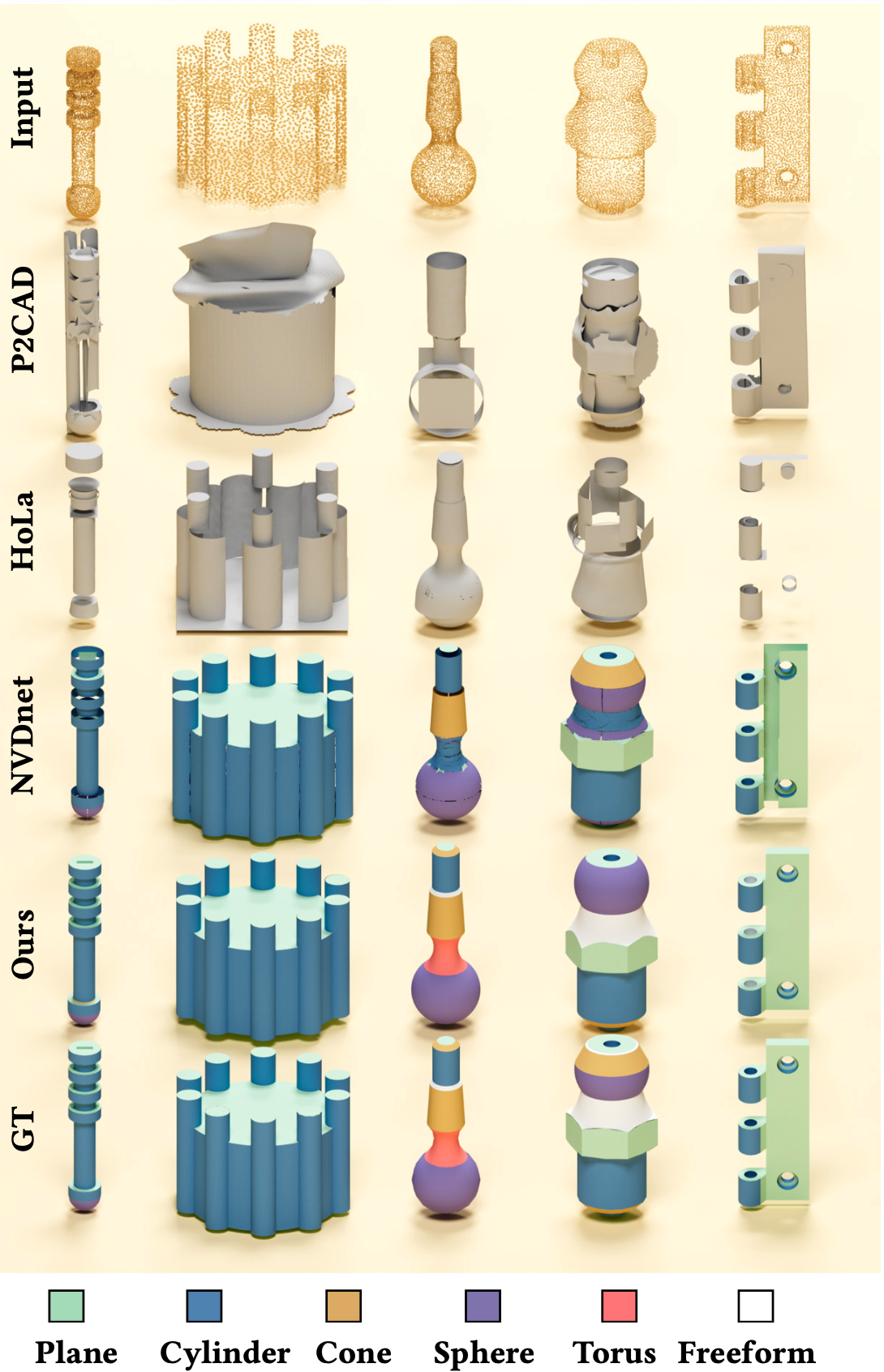}

    \caption{\textbf{Qualitative comparison v.s. baseline methods}. Surfaces are shaded based on predicted types. P2CAD and HoLa fail to recover the shape of the solid and segmentated types. NVDNet misrepresents torus and freeform surfaces by plane, cylinder and sphere. Ours faithfully recovers the geometry and provides meaningful surface semantics.}
    \label{fig:compare}

\end{figure}

Fig.~\ref{fig:compare} qualitatively compares our method with the baselines. P2CAD and NVDNet output meshes, while HoLa and ParaCAD generate B-Rep surfaces. Overall, ParaCAD produces consistently higher-quality reconstructions. In particular, due to limited surface-type support, NVDNet approximates torus and freeform regions with many disconnected planar/cylindrical patches, resulting in fragmented geometry and inconsistent semantics.

\subsubsection{Post-processing validity}
We evaluate the effectiveness of our post-processing pipeline by validating the topological correctness of the generated B-Rep models following HoLa~\cite{liu2025hola}. \edit{We sweep $\lambda$ and $w_b$ over a grid on 50 generated samples and report the Pareto front between geometric fidelity (CD between the sewn B-Rep and the GT point-cloud) and validity ratio, as shown in Tab.~\ref{tab:pareto_lambda_wb}, we select the \textit{Balanced} configuration for the following evaluation.}

\begin{table}[h]
  \centering
  \caption{\edit{Pareto-trimmed configurations from the $\lambda \times w_b$ sweep.}}
  \label{tab:pareto_lambda_wb}
  \begin{tabular}{lcccc}
  \toprule
  Configuration       & $\lambda$ & $w_b$ & CD $\downarrow$ & Validity $\uparrow$ \\
  \midrule
  Best fidelity       & 0.9 & 0.0 & \textbf{0.083} & 39.0\% \\
  \textbf{Balanced} & \textbf{0.6} & \textbf{3.0} & 0.088 & 67.5\% \\
  Best validity    & 0.3 & 5.0 & 0.095 & \textbf{77.5\%} \\
  \bottomrule
  \end{tabular}
  \end{table}
  
\begin{figure}[!htbp]
\centering
\includegraphics[width=0.9\linewidth]{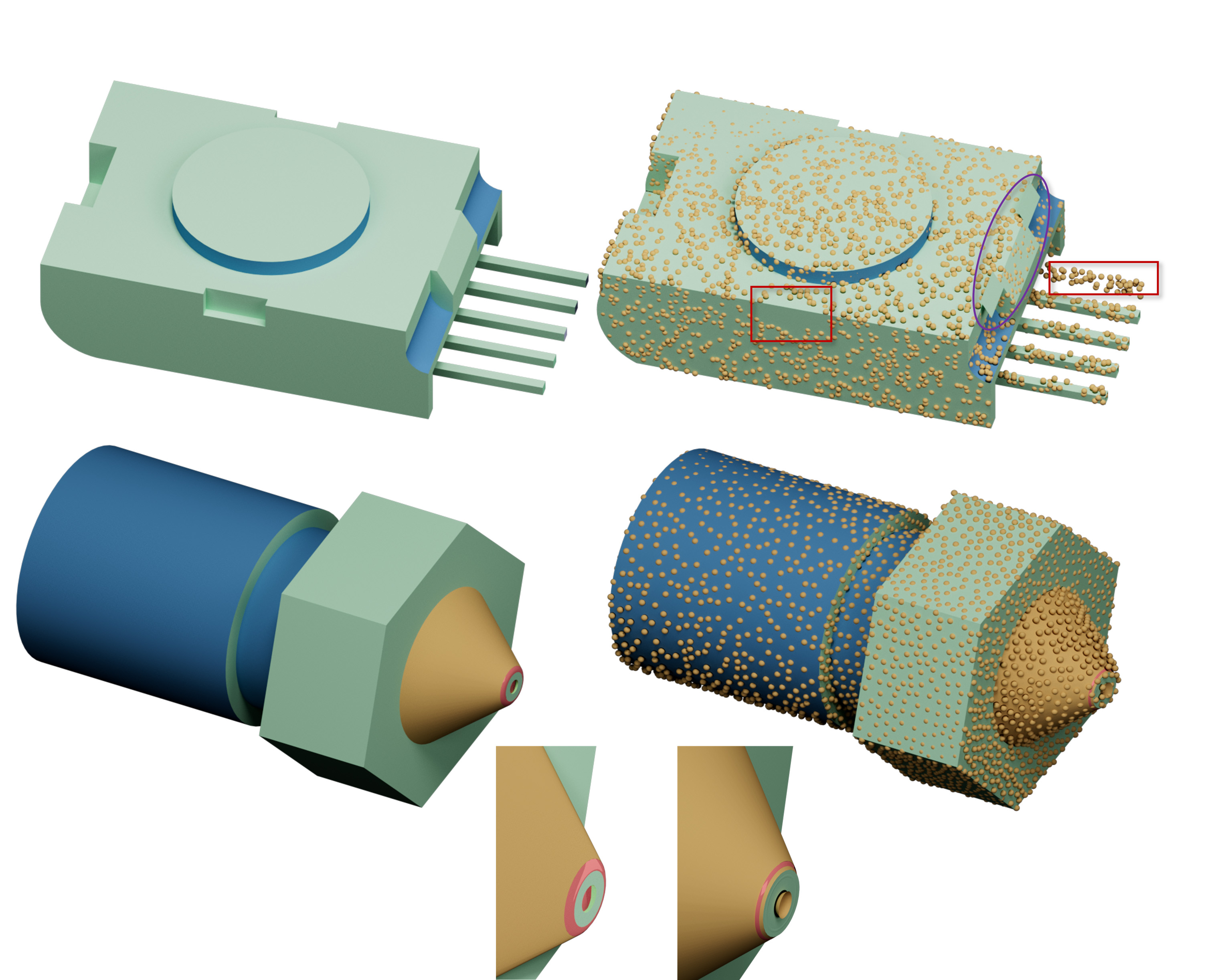}

        \caption{\textbf{Typical failure cases}. GT (Left) vs. Pred (Right). \textbf{Top: } \textit{Red rectangular:} Missing local regions due to limited point-cloud coverage. \textit{Purple ellipse:} Imperfectly cut surfaces become floating structures. \textbf{Bottom:} Hallucination of a cone surface caused by insufficient point cloud resolution.}
        \label{fig:failure}
\end{figure}

We calculate the ratio of valid B-Rep solids among all the generated models to assess the quality of the results. We apply the post-processing method on three set of surface patches and compare their valid ratio: (a) the raw ground truth surface patches. (b) the detokenized ground truth surface patches; (c) the generated surface patches with our autoregressive generator. Tab.~\ref{tab:validity} shows that our post-processing pipeline reliably recovers topologically valid B-Rep solids, achieving high validity on both raw GT patches and detokenized GT patches. Moreover, our point-cloud-conditioned autoregressive generations attain 55.81\% validity after post-processing, substantially outperforming the baseline. \edit{We visualize typical failure cases in Fig.~\ref{fig:failure}.}

\begin{table}[h]
  \centering
  \small
  \setlength{\tabcolsep}{6pt}
  \renewcommand{\arraystretch}{1.2}
    \caption{\textbf{B-Rep validity} on the test set.}
  \label{tab:validity}
  \begin{tabular}{l|c c c c}
    \toprule
    Method &
    HoLa &
    GT &
    Detokenized &
    Generated 
    \\
    \midrule
    Validity (\%)
  & 47.15
  & 77.71
  & 76.60
  & 55.81
  \\
    \bottomrule
  \end{tabular}
\end{table}

\subsection{Ablations}

We ablate the size of the tokenizer codebook by applying our post-processing pipeline to surfaces decoded from quantized tokens. We first collect 691 B-Reps whose ground-truth surface sets can be reliably post-processed into valid solids. For each codebook size, we tokenize these surface sets, decode the surfaces, and run the same post-processing to test whether a valid B-Rep solid is recovered. As shown in

\begin{table}[h]
  \centering
  \small
  \setlength{\tabcolsep}{8pt}
  \renewcommand{\arraystretch}{1.1}
  \caption{\textbf{Ablation on tokenizer codebook size}, reporting the valid solid ratio (\%). We choose 1,024 as our codebook size.}
  \label{tab:ablation_codebook_size}
  \begin{tabular}{c c c c}
    \toprule
    \textbf{Codebook Size} & 2048 & \underline{1024} & 512 \\
    \midrule
    \textbf{Valid Solids (\%)} & 98.38 & 98.08 & 92.78 \\
    \bottomrule
  \end{tabular}
\end{table}

As shown in Tab.~\ref{tab:ablation_codebook_size}, codebook sizes of 2,048 and 1,024 achieve similarly high validity, suggesting both retain sufficient geometric precision, whereas 512 causes a clear drop in valid solids due to quantization-induced loss of surface accuracy.

\subsection{Applications}
\begin{figure}[!htbp]
    \centering
    \includegraphics[width=0.95\linewidth]{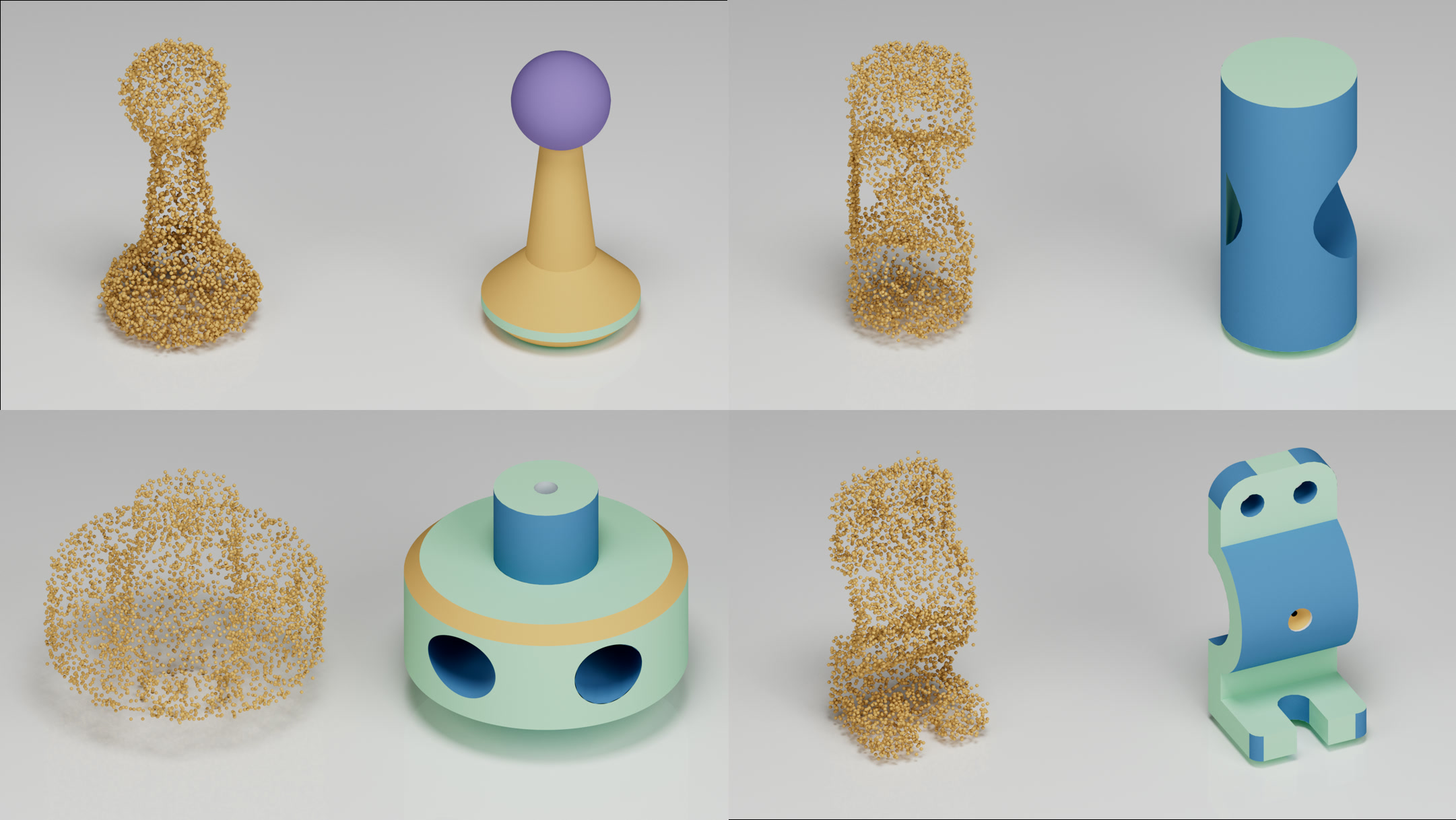}
    \caption{\textbf{Generated B-Reps from point cloud with 2\% Gaussian noise.}}
    \label{fig:noise-pred}
\end{figure}

Our method supports noisy point clouds as input. As shown in Fig.~\ref{fig:noise-pred}, we add 2\% Gaussian noise to both the position and normal of the point cloud, and ParaCAD still manages to produce reasonable B-Rep generation results.

\begin{figure}
    \centering
    \includegraphics[width=0.85\linewidth]{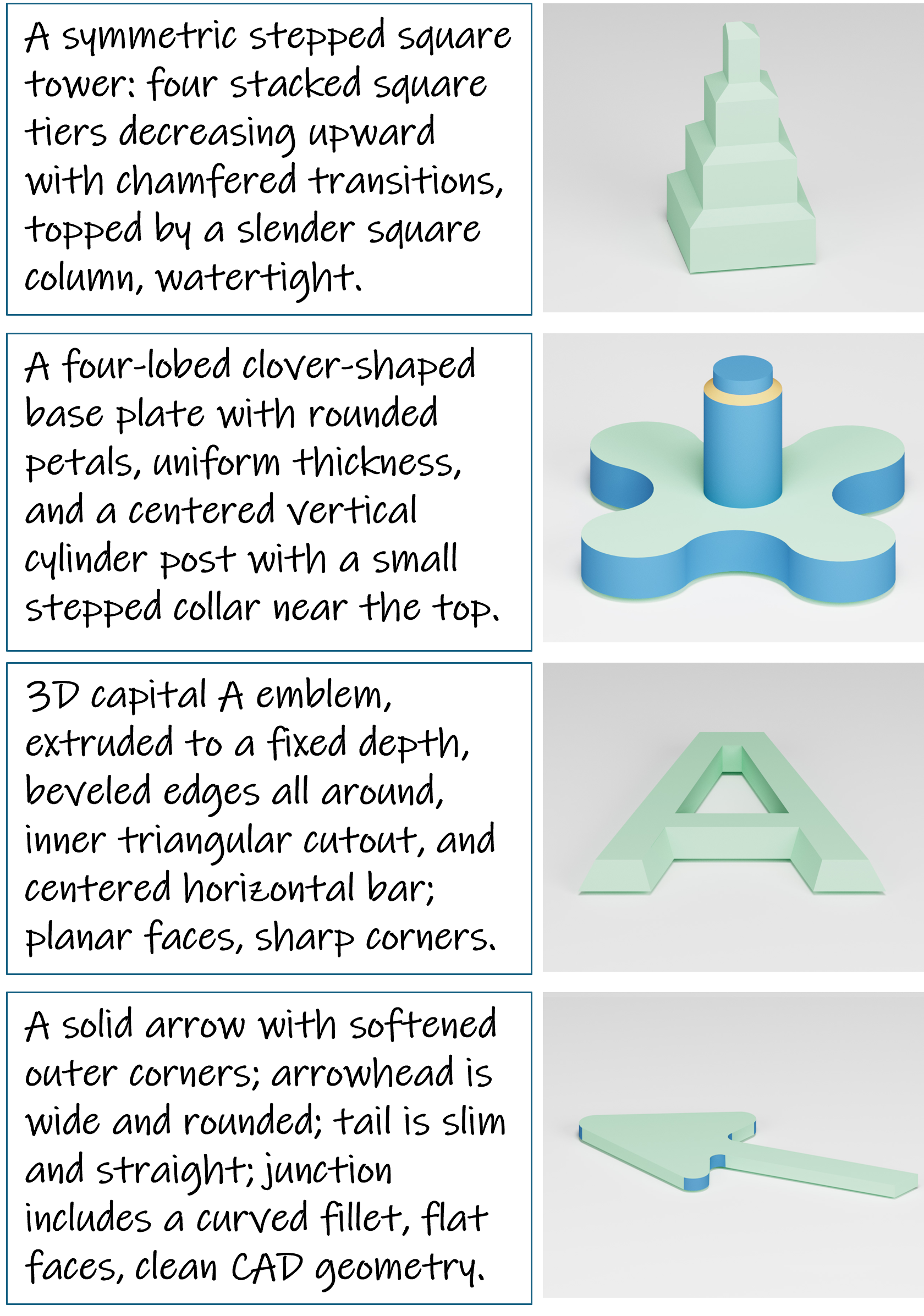}
    \caption{\textbf{Text conditioned B-Rep Generation.} Intermediate point clouds are generated by~\cite{zhang2024clay}.}
    \label{fig:text2cad}
\end{figure}

Using point clouds as a medium, our method can be integrated into existing 3D generation pipelines. For example, as shown in Fig.~\ref{fig:text2cad}, we can generate point clouds from text using a text-to-3D method~\cite{zhang2024clay}, and then obtain the corresponding B-Rep using ParaCAD.

\begin{figure}
    \centering
\includegraphics[width=0.9\linewidth]{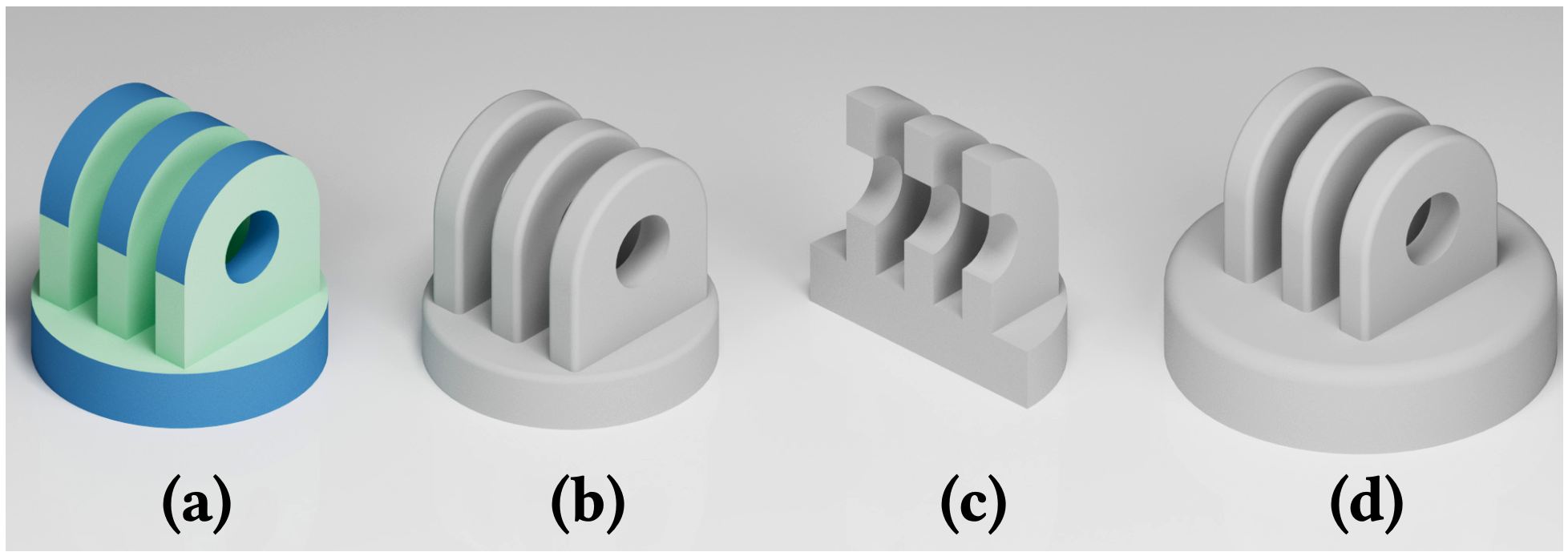}
    \vspace{-2mm}
    \caption{\textbf{Common operations on a generated B-Rep.} (a) Generated B-Rep. (b) Edge fillet. (c) Boolean intersection. (d) Offset of the bottom plate.}
    \label{fig:editing}
\end{figure}

\begin{figure}
    \centering
    \includegraphics[width=0.9\linewidth]{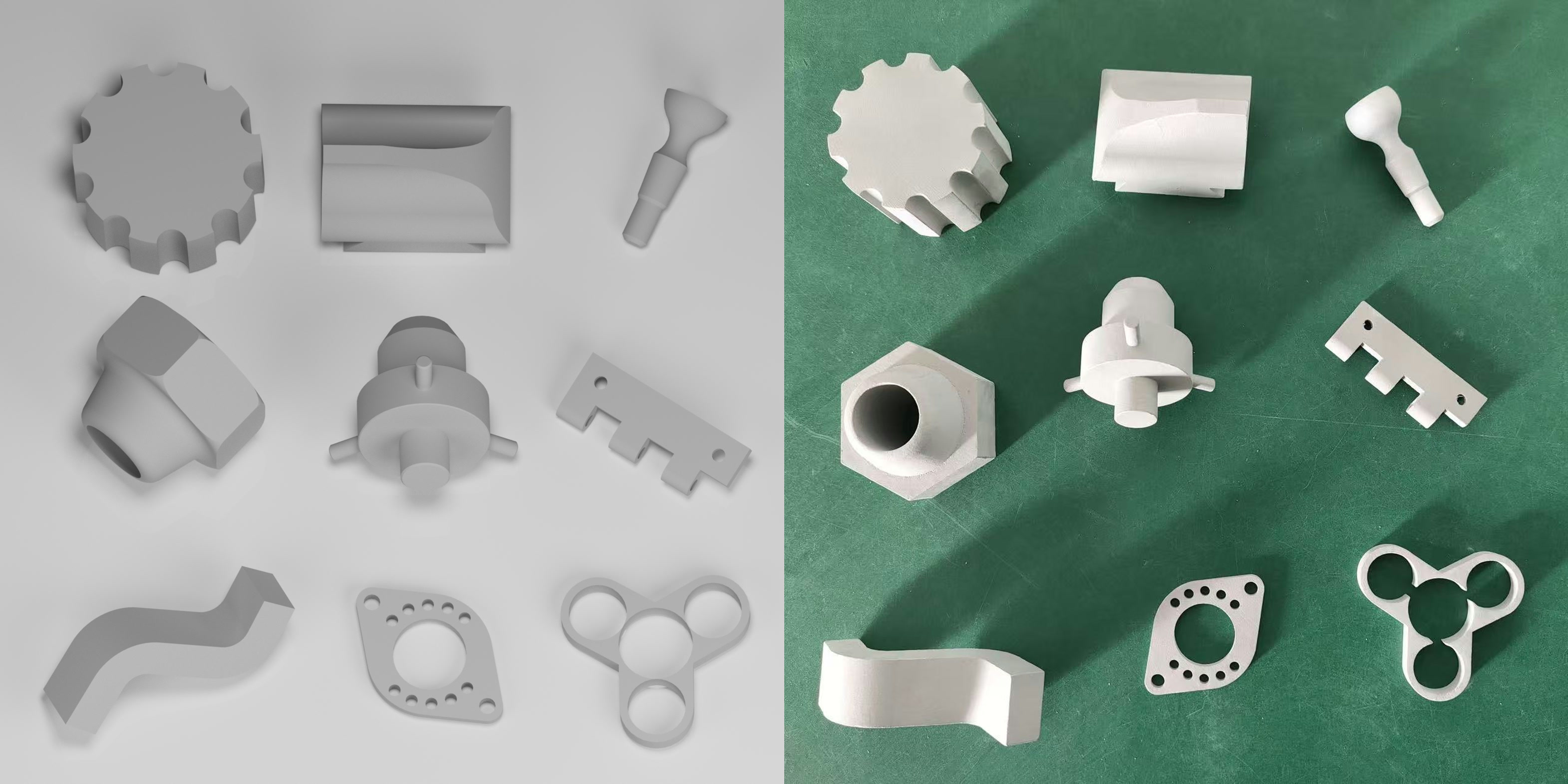}
    \caption{\textbf{Model casted of aluminum alloy}. B-Reps generated by ParaCAD (left) can be directly manufactured (right) using standard mechanical process. }
    \label{fig:build}
\end{figure}

With the help of explicit surface representation, the model we generated can be directly used in downstream tasks. As shown in Fig.~\ref{fig:editing}, common operations like edge fillet, boolean intersection and surface offset on CAD models can be applied to our generated B-Reps. Furthermore, in Fig.~\ref{fig:build} we cast metal models using aluminum alloy based on the generated B-Rep drafts.

\section{Conclusion}
\label{sec:concl}
We introduce ParaCAD, a representation-first approach for CAD generation that operates directly in the native parametric surface space, enabling exact geometric primitives and clean B-Rep generation from point clouds. Our surface-centric formulation simplifies topology recovery and yields high-quality CAD models. ParaCAD provides a convenient tool to streamline the entire process from 3D model generation to industrial design and manufacturing. 

\section{Limitations and Further Works}
\label{sec:limits}
There remains significant room for improvement in achieving higher success rates and more robust, controllable B-Rep generation. Specifically, more accurate UV predictions could enable the post-processing pipeline to handle more complex patch structures. Supporting the generation of higher-order freeform surfaces is also important for practical deployment. Finally, post-training techniques may better align the generative model with user intent and improve watertightness rate.

\begin{acks}
This work was supported in part by the National Natural Science Foundation of China under Grant W2431046, National Key R\&D Program of China 2025YFA1309603, Central Guided Local Science and Technology Foundation of China YDZX20253100001001, and by MoE Key Lab of Intelligent Perceptionand Human-Machine Collaboration (ShanghaiTech University), the Shanghai Frontiers Science Center of Human-centered Artificial Intelligence.
This work was also partially funded by the Innovation and Technology Commission of the HKSAR Government under the ITSP-Platform grants (Ref: ITS/335/23FP and ITS/469/24FP).
\end{acks}

\bibliographystyle{ACM-Reference-Format}
\bibliography{ref}

\clearpage

\begin{figure*}
  \includegraphics[width=\textwidth]{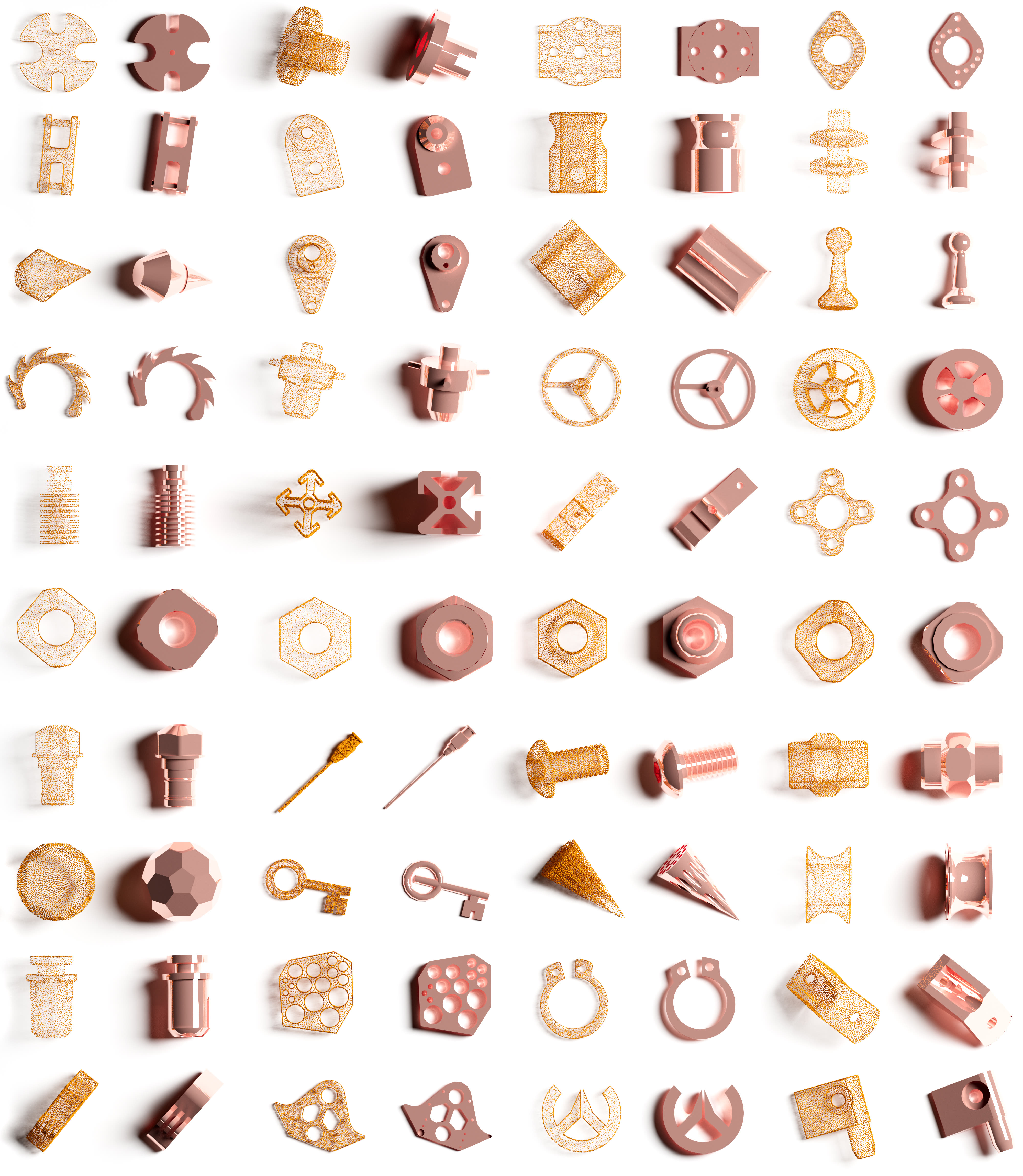}
  \caption{\textbf{From point clouds to manufacturable B-Reps}.
Our method produces native B-Rep models that can be directly rendered with metallic materials and readily integrated into downstream design and manufacturing pipelines. }
  \label{fig:gallery}
\end{figure*}

\end{document}